\documentclass[letterpaper, 10pt, conference]{ieeeconf}  
\IEEEoverridecommandlockouts                              
\usepackage[dvips]{graphics} 
\usepackage{epsfig} 
\usepackage{mathptmx} 
\usepackage{mathtools} 
\usepackage{times} 
\usepackage{amsmath} 
\usepackage{amssymb}  
\usepackage{txfonts}  

\usepackage{algorithm}
\usepackage{algorithmic}
\DeclareMathOperator*{\argmin}{arg\,min}
\DeclareMathOperator*{\argmax}{arg\,max}
\usepackage{subcaption}
\newtheorem{theorem}{Theorem}
\title{\LARGE \bf
Relaxation of the EM Algorithm via Quantum Annealing for Gaussian Mixture Models*
}

\author{Hideyuki Miyahara, Koji Tsumura, and Yuki Sughiyama
\thanks{*This work was supported in part by
  Grant-in-Aid for Scientific Research (B) (25289127), Japan
  Society for the Promotion of Science.}
\thanks{H. Miyahara and K. Tsumura are with
 the Department of Information Physics and Computing, the Graduate School of Information Science and Technology, The University of Tokyo, Japan
  {\tt\small hideyuki\_miyahara@ipc.i.u-tokyo.ac.jp}}%
\thanks{Y. Sughiyama are with
 Institute of Industrial Science, The University of Tokyo, Japan
}
}

\begin{document}

\maketitle
\thispagestyle{empty}
\pagestyle{empty}

\begin{abstract}
We propose a modified expectation-maximization algorithm by introducing the concept of quantum annealing, which we call the deterministic quantum annealing expectation-maximization (DQAEM) algorithm.
The expectation-maximization (EM) algorithm is an established algorithm to compute maximum likelihood estimates and applied to many practical applications.
However, it is known that EM heavily depends on initial values and its estimates are sometimes trapped by local optima.
To solve such a problem, quantum annealing (QA) was proposed as a novel optimization approach motivated by quantum mechanics.
By employing QA, we then formulate DQAEM and present a theorem that supports its stability.
Finally, we demonstrate numerical simulations to confirm its efficiency.
\end{abstract}

\section{Introduction} \label{intro}

Combinatorial optimization is a fundamental issue in both science and engineering.
Although some problems in such optimization can be efficiently solved by well-known algorithms~\cite{Dijkstra01, Kruskal01}, other problems in a class of NP-hard, e.g. the traveling salesman problem, are essentially difficult to solve.

One of the effective approaches for NP-hard problems is simulated annealing (SA), which was proposed by Kirkpatrick \textit{et al.}~\cite{Kirkpatrick01, Kirkpatrick02}.
SA is a generic approach for optimization, in which random numbers that mimic thermal fluctuations are used to go over potential barriers in objective functions.
Furthermore, its global convergence is in some sense guaranteed by Geman and Geman \textit{et al.}~\cite{Geman01}.
After that, a quantum extension of SA, which is called quantum annealing (QA), was proposed in physics~\cite{Apolloni01, Finnila01, Kadowaki01}, and has been intensively studied~\cite{Santoro01, Santoro02, Farhi01, Morita01, Brooke01, Falco01, Falco02, Das01, Marto01}.
In QA, instead of thermal fluctuations, quantum fluctuations are used to overcome potential barriers in objective functions, and it has been reported that QA is more effective than SA for some problems~\cite{Morita01}.
Especially, due to quantum fluctuations, QA exhibits better performance than SA when objective functions have steep multimodality.

Such combinatorial optimization also appears in machine learning, which has attracted much interest recently~\cite{Bishop01, Murphy01}.
For example, some class of data clustering is known to be NP-hard problems~\cite{Aloise01}.
One of common methods for data clustering is as follows.
Assuming data points are generated by Gaussian mixture models (GMMs), we estimate the parameters in GMMs by the expectation-maximization (EM) algorithm~\cite{Dempster01}.
However, parameter estimation sometimes fails since EM depends on initial values and suffers from the problem of local optima.
To relax the problem, Ueda and Nakano proposed a deterministic simulated annealing expectation-maximization (DSAEM) algorithm~\footnote{This algorithm is called the deterministic annealing expectation-maximization algorithm in Ref.~\cite{Ueda01}.}, and it succeeds to relax the difficulty of the multimodality in EM.
This algorithm is based on deterministic simulated annealing (DSA)~\footnote{This algorithm is called deterministic annealing in Ref.~\cite{Rose01}.}, which was proposed by Rose \textit{et al.}~\cite{Rose01, Rose02}.
The essence of these approaches is to make objective functions smooth by introducing thermal fluctuations without random numbers, and the non-convex problem in optimization is considerably managed without increase of numerical cost.

As we have explained, QA is considered to be effective than SA in some conditions~\cite{Morita01}, and thus the quantum version of DSA is expected to be superior to it.
In this paper, we propose a deterministic quantum annealing expectation-maximization (DQAEM) algorithm for Gaussian mixture models because it is expected that quantum fluctuations can relax the problem of local optima in parameter estimation.
In our previous paper~\cite{Miyahara03}, we proposed DQAEM for continuous latent variables, and obtained the result that DQAEM outperformed EM.
However, its applicability is limited because the latent variables are assumed to be continuous and most difficulties in parameter estimation come from optimization of discrete latent variables, such as Gaussian mixture models.
Thus, in this paper, we develop DQAEM for discrete latent variables and apply it to GMMs.
After the formulation of the algorithm, we present a theorem that guarantees its stability.
Finally, to illustrate its efficiency compared to EM, we show numerical simulations, in which DQAEM is applied to GMMs for data clustering.

This paper is organized as follows.
In Sec.~\ref{review-em}, we review GMMs and EM to prepare for DQAEM.
In Sec.~\ref{dqaem-gmm-00}, which is the main section of this paper, we describe the formulation of DQAEM in detail and present a theorem on its convergence.
In Sec.~\ref{sec-numerical-00}, we demonstrate numerical simulations and discuss its efficiency.
In Sec.~\ref{conc}, we conclude this paper.

\section{Review of the expectation-maximization (EM) algorithm and Gaussian mixture models (GMMs) } \label{review-em}

In this section, we review EM to prepare for introducing our DQAEM, and consider an estimation problem of GMMs to formulate DQAEM because it is one of the simplest models with discrete variables.

\subsection{Maximum likelihood estimation (MLE) and the expectation-maximization (EM) algorithm} \label{em-01}

The aim of this subsection is to describe EM because DQAEM is based on it.
First, we review maximum likelihood estimation (MLE) briefly.
Suppose we have $N$ data points $Y_\mathrm{obs} = \{y^{(1)}, y^{(2)}, \dots, y^{(N)}\}$ and they are independent and identically distributed obeying $p(y^{(i)}; \theta)$ where $\theta$ is a parameter.
Moreover we define $p(y^{(i)}, \sigma^{(i)}; \theta)$ as the probability density functions for complete data with the unobservable variables $\{\sigma^{(1)}, \sigma^{(2)}, \dots, \sigma^{(N)}\}$.
Namely, $p(y^{(i)}; \theta) = \sum_{\sigma^{(i)} \in \Omega^{(i)}} p(y^{(i)}, \sigma^{(i)}; \theta)$, where $\Omega^{(i)}$ represents the domain of $\sigma^{(i)}$.
Then the log likelihood function is given by
\begin{align}
\mathcal{L} (Y_\mathrm{obs}; \theta) &= \sum_{i=1}^N \log p(y^{(i)}; \theta) \nonumber \\
                        &= \sum_{i=1}^N \log \sum_{\sigma^{(i)} \in \Omega^{(i)}} p(y^{(i)}, \sigma^{(i)}; \theta). \label{log-likelihood-GMM-01}
\end{align}
Note that $i$ in $y^{(i)}$ and $\sigma^{(i)}$ is the index for each observed data point.
MLE is a technique to estimate the parameter $\theta$ in model distributions that maximize the log likelihood function $\mathcal{L}(Y_\mathrm{obs}; \theta)$.

In general, maximizing the log likelihood function $\mathcal{L} (Y_\mathrm{obs}; \theta)$ with respect to $\theta$ is difficult because it is sometimes a non-convex optimization, and then we replace it with its lower bound.
Using Jensen's inequality, we have the following inequality
\begin{align}
\mathcal{L}(Y_\mathrm{obs};\theta) &= \sum_{i=1}^N \log \sum_{\sigma^{(i)} \in \Omega^{(i)}} P(\sigma^{(i)}| y^{(i)}; \theta') \frac{p(y^{(i)}, \sigma^{(i)}; \theta)}{P(\sigma^{(i)}| y^{(i)}; \theta')} \nonumber \\
                                   &\ge \sum_{i=1}^N \sum_{\sigma^{(i)} \in \Omega^{(i)}} P(\sigma^{(i)}| y^{(i)}; \theta') \log \frac{p(y^{(i)}, \sigma^{(i)}; \theta)}{P(\sigma^{(i)}| y^{(i)}; \theta')} \nonumber \\
                                   &\ge Q(\theta;\theta') \nonumber \\
                                   & \quad - \sum_{i=1}^N \sum_{\sigma^{(i)} \in \Omega^{(i)}} P(\sigma^{(i)}|y^{(i)};\theta') \log P(\sigma^{(i)}|y^{(i)};\theta'), \nonumber \\
Q(\theta;\theta') &= \sum_{i=1}^N \sum_{\sigma^{(i)} \in \Omega^{(i)}} P(\sigma^{(i)}| y^{(i)}; \theta') \log p(y^{(i)}, \sigma^{(i)}; \theta), \label{EM-Mstep01}
\end{align}
where $\theta'$ is an arbitrary parameter and $P(\sigma^{(i)}| y^{(i)}; \theta')$ is the conditional probability.
Then, the procedure of EM consists of the following two steps.
The first one, which is called the E step, is to compute the conditional probability $P(\sigma^{(i)}| y^{(i)}; \theta)$ by
\begin{align}
P(\sigma^{(i)}| y^{(i)}; \theta') &= \frac{p(y^{(i)}, \sigma^{(i)}; \theta')}{p(y^{(i)}; \theta')}, \label{EM-Estep01} \\
p(y^{(i)}; \theta') &= \sum_{\sigma^{(i)} \in \Omega^{(i)}} p(y^{(i)}, \sigma^{(i)}; \theta'). \nonumber
\end{align}
Here we have used Bayes' rule.
The second one, which is called the M step, is to maximize the $Q$ function~\eqref{EM-Mstep01} with respect to $\theta$ instead of $\mathcal{L} (Y_\mathrm{obs}; \theta)$.
Denoting the tentative estimated parameter at the $t$-th iteration by $\theta^{(t)}$, the estimated parameter is updated by
\begin{align}
\theta^{(t+1)} = \argmax_\theta Q(\theta; \theta^{(t)}). \nonumber
\end{align}
At the end of this subsection, we summarize EM in Algo.~\ref{EM-algorithm-01}.
\begin{algorithm}[t]
\caption{Expectation-maximization (EM) algorithm}
\label{EM-algorithm-01}
\begin{algorithmic}[1]
\STATE initialize $\theta^{(0)}$ and set $t \leftarrow 0$
\WHILE{convergence criterion is satisfied}
\STATE calculate $P(\sigma^{(i)}| y^{(i)}; \theta^{(t)}) \ (i = 1, \dots, N)$ with~\eqref{EM-Estep01} \ (E step)
\STATE calculate $\theta^{(t+1)} = \argmax_\theta Q(\theta;\theta^{(t)})$ where $Q(\theta;\theta^{(t)})$ is~\eqref{EM-Mstep01} \ (M step)
\ENDWHILE
\end{algorithmic}
\end{algorithm}

\subsection{Gaussian mixture models (GMMs)} \label{gmm-01}

Here we introduce GMMs and its quantum mechanical representation.
We follow the notations in Refs.~\cite{Bishop01, Murphy01}.
Let $y$ and $\sigma$ denote continuous observable and discrete unobservable variables.
Here, we assume that $\Omega$, which is the domain of $\sigma$, is given by $\{1_k\}_{k=1}^K$, where
\begin{align}
1_k &= [\underbrace{0, \dots, 0}_{k - 1}, 1, \underbrace{0, \dots, 0}_{K - k}]^\intercal, \nonumber
\end{align}
for $k = 1, \dots, K$, and then the number of elements in $\Omega$ is $K$.
Specifically, $\sigma = 1_k$ when $\sigma$ denotes the $k$-th element in $\Omega$.

Using the above notation, the probability density function of GMMs is given by
\begin{align}
p(y; \theta) &= \sum_{k=1}^K p(y| \sigma = 1_k; \theta) P(\sigma = 1_k; \theta), \nonumber
\end{align}
where
\begin{align}
p(y| \sigma = 1_{k}; \theta) &= g(y; \mu_k, \Sigma_k), \nonumber \\
P(\sigma = 1_k; \theta) &= \pi_k \ (k = 1,\dots, K), \nonumber
\end{align}
$\{\pi_k\}_{k=1}^K$ satisfies $\sum_k \pi_k = 1$, $g(y; \mu_k, \Sigma_k)$ is a Gaussian function with mean $\mu_k$ and covariance $\Sigma_k$ for $k = 1, \dots, K$, and $\theta = \{\pi_k, \mu_k, \Sigma_k\}_{k=1}^K$.
The joint probability density function for GMMs is therefore given by
\begin{align}
p(y, \sigma; \theta) &= \prod_k \left[ p(y| \sigma = 1_{k}; \mu_k, \Sigma_k) p(\sigma = 1_{k}; \pi_k) \right]^{\sigma_{k}} \nonumber \\
             &= \prod_k \left[ \pi_k g(y; \mu_k, \Sigma_k) \right]^{\sigma_{k}}, \label{joint-GMM-01}
\end{align}
where $\sigma_k$ is the $k$-th element of $\sigma$.

To introduce quantum fluctuations, we need to rewrite the above equations in the Hamiltonian formulation.
Taking the logarithm of~\eqref{joint-GMM-01}, the Hamiltonian for GMMs can then be written as
\begin{align}
H(y, \sigma; \theta) &= \sum_k h_k \sigma_{k}, \label{hamil-GMM-01}
\end{align}
where $h_k = - \log \{ \pi_k g(y; \mu_k, \Sigma_k)\}$ for $k = 1, \dots, K$.
Here, we introduce ket vectors, bra vectors and ``spin" operators to rewrite \eqref{hamil-GMM-01} in the manner of quantum mechanics.
First, we define the ket vector $|\sigma = 1_k \rangle$ by $1_k$ and the ``spin" operator $\hat{\sigma}_k$ by
\begin{align}
\hat{\sigma}_k &= | \sigma = 1_k \rangle \langle \sigma = 1_k | \nonumber \\
               &= \mathrm{diag} (\underbrace{0, \dots, 0}_{k - 1}, 1, \underbrace{0, \dots, 0}_{K - k}), \nonumber
\end{align}
respectively, where the bra vector $\langle \sigma = 1_j |$ satisfies the orthonormal condition $\langle \sigma = 1_j | \sigma = 1_i \rangle = \delta_{ij}$.
Replacing $\sigma$ with $\hat{\sigma}$, we have the Hamiltonian operator
\begin{align}
H(y, \hat{\sigma}; \theta) &= \sum_k h_k \hat{\sigma}_{k}, \nonumber \\
                           &= \mathrm{diag}(h_{1}, h_{2}, \dots, h_{K}), \label{original-hamil-03}
\end{align}
and this satisfies
\begin{align}
\langle \sigma = 1_i | H(y, \hat{\sigma}; \theta) | \sigma = 1_j \rangle &=  h_i \delta_{ij}, \nonumber
\end{align}
where $\delta_{ij}$ is the Kronecker delta.
We use this formulation to describe DQAEM in the following section.
Note that a similar expression is presented in Ref.~\cite{Tanaka02}.

\section{Deterministic quantum annealing expectation-maximization algorithm (DQAEM)} \label{dqaem-gmm-00}

First, we formulate DQAEM by using the quantum representation described in the previous section.
Then we discuss its stability by showing the monotonicity of the free energy during the algorithm.

\subsection{Formulation} \label{dqaem-gmm-01}

In this subsection, we formulate DQAEM by employing the concept of quantum annealing~\cite{Kadowaki01} (also see App.~\ref{app-qa-01}).
First, we rewrite EM in the quantum representation.
The log likelihood function \eqref{log-likelihood-GMM-01} is rewritten as
\begin{align}
\mathcal{L} (Y_\mathrm{obs}; \theta) &= \sum_{i=1}^N \log \mathrm{Tr} \left[ p(y^{(i)}, \hat{\sigma}^{(i)}; \theta) \right]. \label{log-likelihood-51-01}
\end{align}
Note that $\mathrm{Tr} \left[ \cdot \right] = \sum_{k} \langle \sigma^{(i)} = 1_k | \left[ \cdot \right] | \sigma^{(i)} = 1_k \rangle$.
As we have explained in Sec.~\ref{em-01}, the $Q$ function~\eqref{EM-Mstep01} is maximized in the M step of EM.
Similarly to \eqref{log-likelihood-51-01}, the quantum representation of the $Q$ function~\eqref{EM-Mstep01} is given by
\begin{align}
Q &(\theta; \theta') = \sum_{i=1}^N \mathrm{Tr} \left[ P(\hat{\sigma}^{(i)}| y^{(i)}; \theta') \log p (y^{(i)}, \hat{\sigma}^{(i)}; \theta) \right], \nonumber
\end{align}
where
\begin{align}
p (y^{(i)}, \hat{\sigma}^{(i)}; \theta) &= \exp \{ - (H(y^{(i)}, \hat{\sigma}^{(i)}; \theta))\}, \label{dist-46}
\end{align}
and $H(y^{(i)}, \hat{\sigma}^{(i)}; \theta)$ is in Eq.~\eqref{original-hamil-03}.
Furthermore, the conditional probability $P (\hat{\sigma}^{(i)}|y^{(i)}; \theta)$ is computed using Bayes' rule.
That is,
\begin{align}
P (\hat{\sigma}^{(i)}|y^{(i)}; \theta) &= \frac{p (y^{(i)}, \hat{\sigma}^{(i)}; \theta)}{\mathcal{Z}^{(i)} (\theta)}. \nonumber
\end{align}
Here, the normalization factor, which is called the partition function in physics, has the form
\begin{align}
\mathcal{Z}^{(i)} (\theta) &= \mathrm{Tr} \left[ p (y^{(i)}, \hat{\sigma}^{(i)}; \theta) \right]. \nonumber
\end{align}

Now we begin to formulate DQAEM.
To introduce quantum fluctuations, we add $H' (\Gamma) = \Gamma \hat{\sigma}'$ whose $\hat{\sigma}'$ satisfies $[\hat{\sigma}_k, \hat{\sigma}'] \ne 0$ for $k = 1, \dots, K$ to the original Hamiltonian $H$, and then \eqref{dist-46} is converted to
\begin{align}
p_{\Gamma} (y^{(i)}, \hat{\sigma}^{(i)}; \theta) &= \exp \{ - (H(y^{(i)}, \hat{\sigma}^{(i)}; \theta) + H'(\Gamma) )\}. \label{quantum-dist-01}
\end{align}
In MLE, the log likelihood function~\eqref{log-likelihood-51-01} is optimized.
On the other hand, the objective function in DQAEM, which is called the free energy, is given by
\begin{align}
F_{\Gamma} (\theta) = - \log \mathcal{Z}_{\Gamma} (\theta), \nonumber
\end{align}
where
\begin{align}
\mathcal{Z}_{\Gamma} (\theta) &= \prod_{i=1}^N \mathcal{Z}_{\Gamma}^{(i)} (\theta), \nonumber \\
\mathcal{Z}_{\Gamma}^{(i)} (\theta) &= \mathrm{Tr} \left[ p_{\Gamma} (y^{(i)}, \hat{\sigma}^{(i)}; \theta) \right]. \nonumber
\end{align}
By taking in into account $H'(0)=0$ and comparing to Eq.~\eqref{log-likelihood-51-01}, we obtain the relation between the free energy and the log likelihood function as
\begin{align}
F_{\Gamma = 0} (\theta) = - \mathcal{L}(Y_\mathrm{obs}; \theta). \label{identity-01}
\end{align}
Thus we can say that the negative free energy at $\Gamma = 0$ is the log likelihood function.

Next, we define the function $U_{\Gamma} (\theta; \theta')$ to formulate DQAEM, which corresponds to the $Q$ function in EM.
Using \eqref{quantum-dist-01}, the function $U_{\Gamma} (\theta; \theta')$ has the form
\begin{align}
U_{\Gamma} &(\theta; \theta') = \sum_{i=1}^N \mathrm{Tr} \left[ P_{\Gamma} (\hat{\sigma}^{(i)}| y^{(i)}; \theta') \log p_{\Gamma} (y^{(i)}, \hat{\sigma}^{(i)}; \theta) \right], \label{QAEM-Mstep01}
\end{align}
where
\begin{align}
P_{\Gamma} (\hat{\sigma}^{(i)}|y^{(i)}; \theta) &= \frac{p_{\Gamma} (y^{(i)}, \hat{\sigma}^{(i)}; \theta)}{\mathcal{Z}_{\Gamma}^{(i)} (\theta)}. \label{posterior02}
\end{align}

Then DQAEM is composed of the following two steps.
The first one is to compute the conditional probability~\eqref{posterior02}, and this is called the E step of DQAEM.
The second one is to update the parameter $\theta^{(t)}$ by minimizing the function $U_{\Gamma} (\theta; \theta')$~\eqref{QAEM-Mstep01}.
That is,
\begin{align}
\theta^{(t+1)} = \argmin_\theta U_{\Gamma} (\theta, \theta^{(t)}), \nonumber
\end{align}
and this is called the M step in DQAEM.
Furthermore, we decrease $\Gamma$ during the iterations.
We summarize DQAEM in Algo.~\ref{DQAEM-algorithm-01}.
\begin{algorithm}[t]
\caption{Deterministic quantum annealing expectation-maximization (DQAEM) algorithm}
\label{DQAEM-algorithm-01}
\begin{algorithmic}[1]
\STATE set $\Gamma \leftarrow \Gamma_\mathrm{init}$
\STATE initialize $\theta^{(0)}$ and set $t \leftarrow 0$
\WHILE{convergence criteria is satisfied}
\STATE calculate $P_{\Gamma} (\hat{\sigma}^{(i)}| y^{(i)}; \theta^{(t)}) \ (i = 1, \dots, N)$ with~\eqref{posterior02} \ (E step)
\STATE calculate $\theta^{(t+1)} = \argmin_{\theta} U_{\Gamma}(\theta;\theta^{(t)})$ with~\eqref{QAEM-Mstep01} \ (M step)
\STATE decrease $\Gamma$
\ENDWHILE
\end{algorithmic}
\end{algorithm}

\subsection{Convergence theorem} \label{dqaem-gmm-02}

We have proposed DQAEM in the previous subsection.
Here, we present the theorem that guarantees its stability via iterations.
\begin{theorem} \label{theorem01}
Let $\theta^{(t+1)} = \argmin_\theta U_{\Gamma}(\theta;\theta^{(t)})$.
Then $F_{\Gamma}(\theta^{(t+1)}) \le F_{\Gamma}(\theta^{(t)})$ holds.
Moreover, the equality holds if and only if $U_{\Gamma}(\theta^{(t+1)}; \theta^{(t)}) = U_{\Gamma}(\theta^{(t)};\theta^{(t)})$ and $S_{\Gamma}(\theta^{(t+1)};\theta^{(t)}) = S_{\Gamma}(\theta^{(t)};\theta^{(t)})$,
where $S_{\Gamma}(\theta;\theta') = \sum_{i=1}^N \mathrm{Tr} \left[ P_{\Gamma} (\hat{\sigma}^{(i)}| y^{(i)}; \theta') \log P_{\Gamma} (\hat{\sigma}^{(i)}| y^{(i)} ; \theta) \right]$.
\end{theorem}

This theorem insists that DQAEM converges at least the global optimum or a local optimum.
We mention that the global convergence of EM is discussed by Dempster \textit{et al.}~\cite{Dempster01} and Wu~\cite{Wu01}, and their discussions apply to DQAEM.

\section{Numerical simulations} \label{sec-numerical-00}

In this section, we carry out numerical simulations to confirm the performance of DQAEM.
In the first subsection, we present the setup of numerical simulations, and, in the following subsection, we provide numerical results.

\subsection{Mathematical setup} \label{sec-numerical-01}

We estimate the parameters of GMMs by using both DQAEM and EM.
Suppose $N$ data points $Y_\mathrm{obs} = \{y^{(1)}, y^{(2)}, \dots, y^{(N)}\}$ are identically sampled by GMMs with $K = 3$.
Here, a GMM is given by~\eqref{joint-GMM-01}.
In EM, the updating equations for $\theta = \{\pi_k, \mu_k, \Sigma_k \}_{k=1}^K$ are determined by the derivative of the $Q$ function~\eqref{EM-Mstep01} with respect to $\theta$.
The parameter $\theta^{(t+1)}$ of GMMs at the $t+1$-th iteration is then given by
\begin{align}
\pi_k^{(t+1)} &= \frac{1}{N} \sum_{i=1}^N P(\sigma^{(i)} = 1_k | y^{(i)}; \theta^{(t)}), \label{update-EM-01} \\
\mu_k^{(t+1)} &= \frac{\sum_{i=1}^N y^{(i)} P(\sigma^{(i)} = 1_k| y^{(i)}; \theta^{(t)})}{\sum_{i=1}^N P(\sigma^{(i)} = 1_k| y^{(i)}; \theta^{(t)})}, \label{update-EM-02} \\
\Sigma_{k}^{(t+1)} &= \frac{\sum_{i=1}^N (y^{(i)} - \mu_k^{(t+1)}) (y^{(i)} - \mu_k^{(t+1)})^\intercal P(\sigma^{(i)} = 1_k| y^{(i)}; \theta^{(t)})}{\sum_{i=1}^N P(\sigma^{(i)} = 1_k| y^{(i)}; \theta^{(t)})}, \label{update-EM-03}
\end{align}
where $\theta^{(t)}$ is the tentative estimated parameter at the $t$-th iteration.

In DQAEM, the updating equations for $\theta$ are determined by the derivative of the function $U_{\Gamma} (\theta, \theta')$ in~\eqref{QAEM-Mstep01} with respect to $\theta$, and then $P(\sigma^{(i)} = 1_k | y^{(i)}; \theta^{(t)})$ in~\eqref{update-EM-01}, \eqref{update-EM-02} and \eqref{update-EM-03} are replaced by $P_\mathrm{QA}(\sigma^{(i)} = 1_k| y^{(i)}; \theta^{(t)}) = \langle \sigma^{(i)} = 1_k | P_{\Gamma} (\hat{\sigma}^{(i)}| y^{(i)}; \theta^{(t)}) | \sigma^{(i)} = 1_k \rangle$.
That is, the updating equations for DQAEM are given by
\begin{align}
\pi_k^{(t+1)} &= \frac{1}{N} \sum_{i=1}^N P_\mathrm{QA}(\sigma^{(i)} = 1_k| y^{(i)}; \theta^{(t)}), \nonumber \\
\mu_k^{(t+1)} &= \frac{\sum_{i=1}^N y^{(i)} P_\mathrm{QA}(\sigma^{(i)} = 1_k| y^{(i)}; \theta^{(t)})}{\sum_{i=1}^N P_\mathrm{QA}(\sigma^{(i)} = 1_k| y^{(i)}; \theta^{(t)})}, \nonumber \\
\Sigma_{k}^{(t+1)} &= \frac{\sum_{i=1}^N (y^{(i)} - \mu_k^{(t+1)}) (y^{(i)} - \mu_k^{(t+1)})^\intercal P_\mathrm{QA}(\sigma^{(i)} = 1_k| y^{(i)}; \theta^{(t)})}{\sum_{i=1}^N P_\mathrm{QA}(\sigma^{(i)} = 1_k| y^{(i)}; \theta^{(t)})}. \nonumber
\end{align}
Note that the quantum effects for parameter estimation comes from $P_\mathrm{QA}(\sigma^{(i)} = 1_k| y^{(i)}; \theta^{(t)})$.
The annealing parameter $\Gamma$ are varied from initial values to $0$ via iterations.

In this section, assume that, in matrix notation, $\hat{\sigma}'$ is given by
\begin{align}
\sigma'=
\begin{bmatrix}
0 & 1 & 1 \\
1 & 0 & 1 \\
1 & 1 & 0
\end{bmatrix}. \nonumber
\end{align}
Obviously $[\hat{\sigma}_k, \hat{\sigma}'] \ne 0$ is satisfied.
Note that the size of the Hamiltonian is determined by assumed number of mixtures.

\subsection{Numerical results} \label{sec-numerical-02}

\begin{figure}[t]
\begin{subfigure}[t]{0.5\textwidth}
\centering
\includegraphics[scale=0.65]{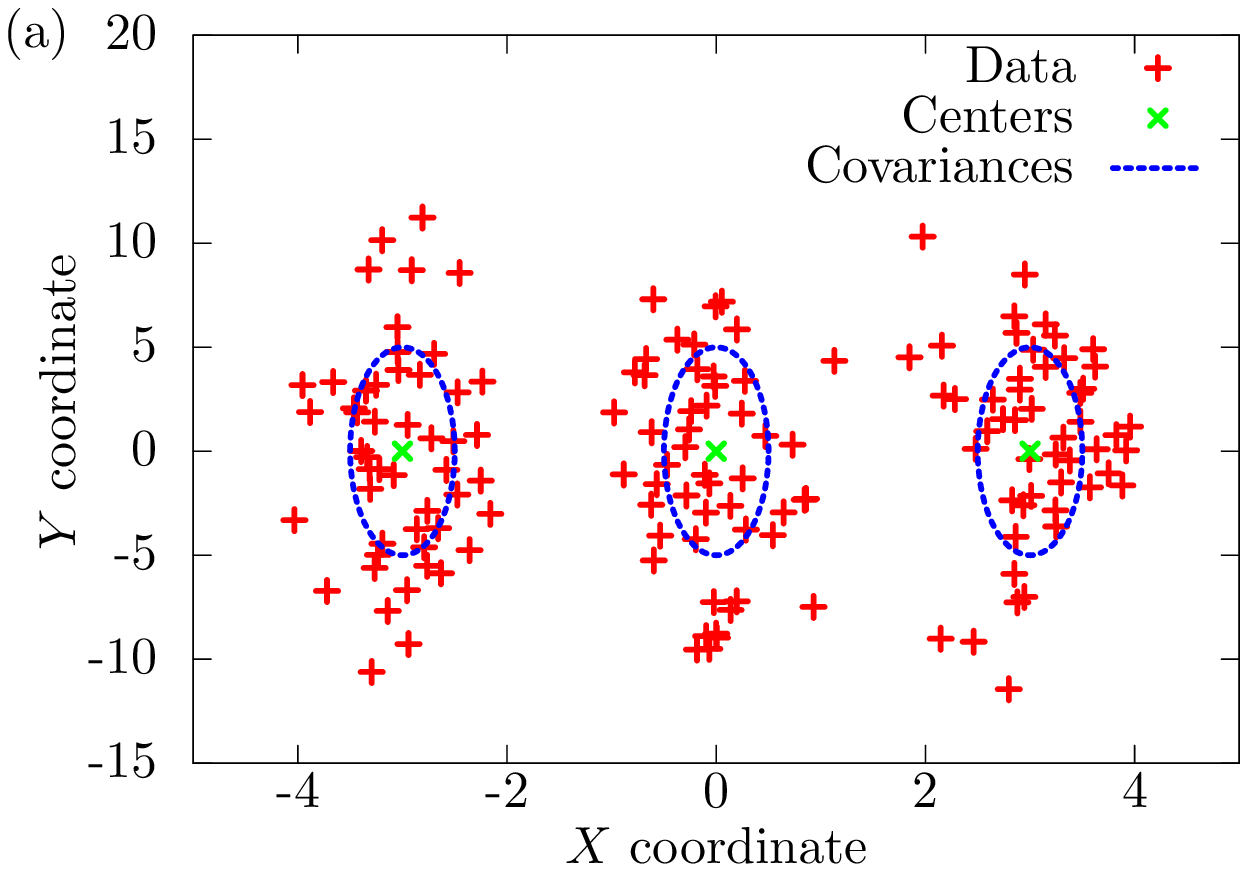}
\end{subfigure}
\begin{subfigure}[t]{0.5\textwidth}
\centering
\includegraphics[scale=0.65]{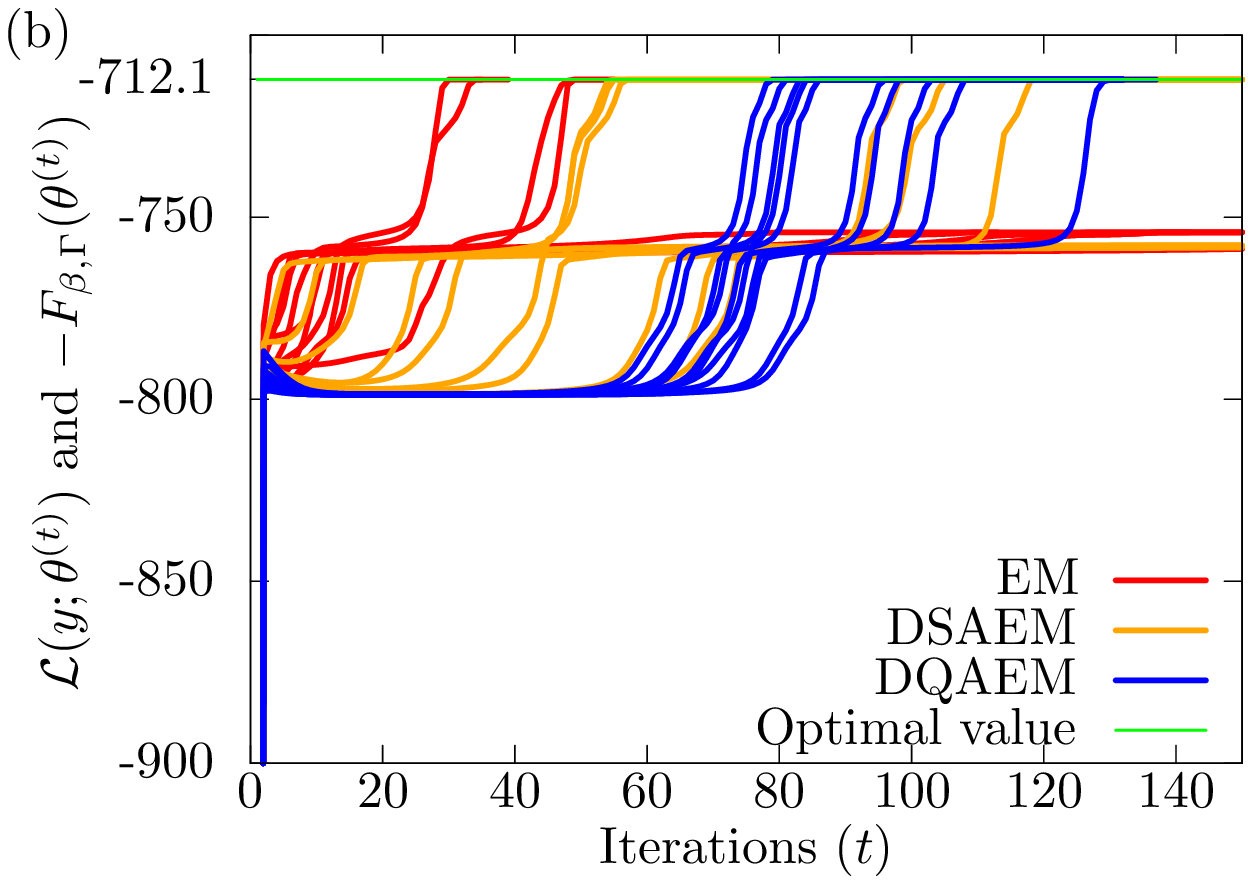}
\end{subfigure}
\caption{(a) Data set generated by three Gaussian functions whose means are $(X, Y) = (-3, 0)$, $(0, 0)$ and $(3,0)$. (b) Number of iterations (log scale) vs typical transitions of the log likelihood functions in EM and the negative free energies in DSAEM and DQAEM.}
\label{numerical-01-01}
\end{figure}

In this subsection, using the data set shown in Fig.~\ref{numerical-01-01}(a), we compare DQAEM, EM, and DSAEM, which was proposed in Ref.~\cite{Ueda01}.
This data set is generated by the GMM that consists of three two-dimensional Gaussian functions whose means are $(X, Y) = (-3, 0)$, $(0, 0)$ and $(3, 0)$.
Here we set $\Gamma_\mathrm{init} = 1.0$ in DQAEM to discuss the effect of quantum fluctuations simply.
We also choose the annealing parameter in DSAEM as $\beta_\mathrm{init} = 0.7$.
Note that, in DSAEM, the annealing parameter is given by temperature.
Furthermore, we exponentially vary $\beta$ and $\Gamma$ to 1 and 0, respectively.
We plot transitions of the log likelihood functions of EM and the negative free energies of DSAEM and DQAEM in Fig.~\ref{numerical-01-01}(b) by red lines, orange lines, and blue lines, respectively.
The value of $-712.1$ depicted by the green line in Fig.~\ref{numerical-01-01}(b) is the optimal value in these numerical simulations.
DQAEM, EM, and DSAEM give the optimal estimate or suboptimal estimates depending on initial optimization values.

\begin{figure}[t]
\begin{subfigure}[t]{0.5\textwidth}
\centering
\includegraphics[scale=0.65]{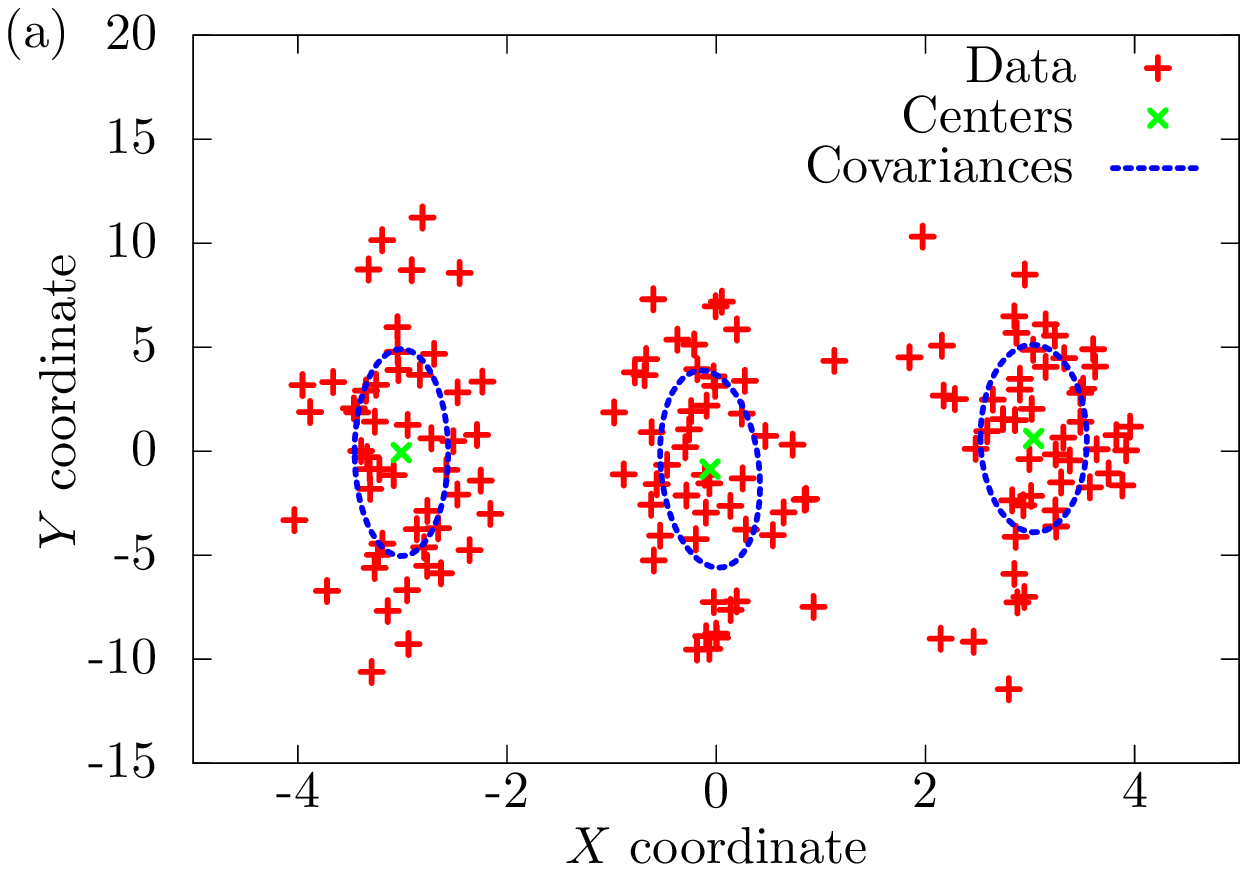}
\end{subfigure}
\begin{subfigure}[t]{0.5\textwidth}
\centering
\includegraphics[scale=0.65]{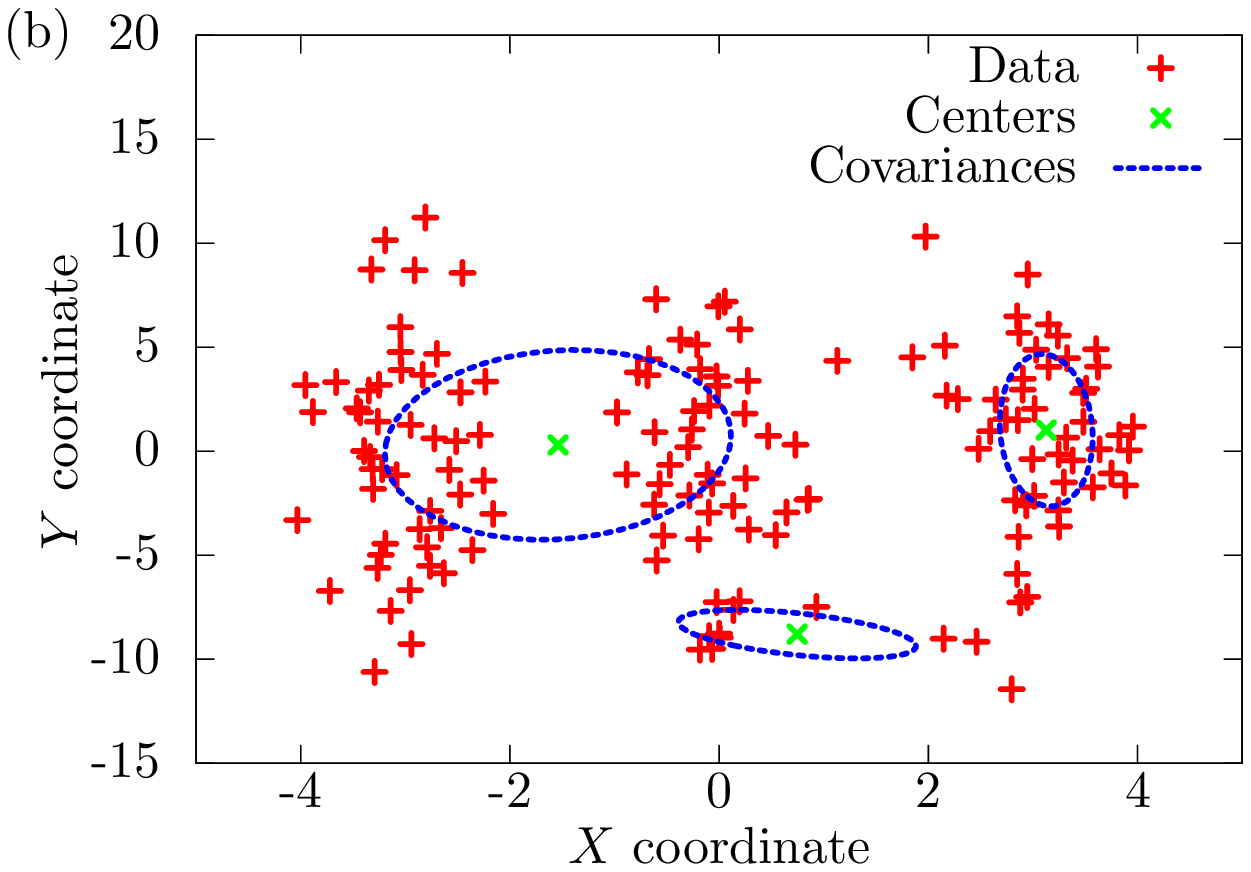}
\end{subfigure}
\caption{Estimated Gaussian functions (a) in the case where the log likelihood function is the value of $-712.1$ and (b) in one of the cases where the log likelihood function is lower than the optimal value. Green crosses and blue lines represent the estimated means and covariances, respectively.}
\label{numerical-02-01}
\end{figure}

To understand visually how DQAEM and EM behave in parameter estimation, we illustrate estimated Gaussian functions in the case where the log likelihood function is $-712.1$ in Fig.~\ref{numerical-02-01}(a) and in one of the cases where the log likelihood function is lower than the optimal value in Fig.~\ref{numerical-02-01}(b).
The case demonstrated in Fig.~\ref{numerical-02-01}(b) clearly fails in data clustering.

However, the ratios of success for DQAEM, EM, and DSAEM are much different.
To see the ratios of success and failure for DQAEM and EM, we performed DQAEM and EM with same initial optimization values $1000$ times, respectively, and summarized the results in Table~\ref{joint-table-01}.
\begin{table}[t]
\caption{Ratios of success and failure for DQAEM and EM.}
\label{joint-table-01}
\begin{center}
\begin{tabular}{|c|c|p{5em}|p{5em}|p{5em}|}
\hline
\multicolumn{2}{|c|}{}  & \multicolumn{3}{c|}{DQAEM}   \\ \cline{3-5}
\multicolumn{2}{|c|}{}  & Success & Fail    & Total      \\ \hline
                        & Success     & 55.9 \% &  0.7 \% & 56.6  \% \\ \cline{2-5}
EM                      & Fail        & 41.5 \% &  1.9 \% & 43.4  \% \\ \cline{2-5}
                        & Total       & 97.4 \% &  2.6 \% & 100.0 \% \\ \hline
\end{tabular}
\end{center}
\end{table}
Here, we have defined the ``success" of DQAEM and EM when square errors between the estimated means of three Gaussian functions and the true means are less than $0.3$ times the covariances of three Gaussian functions.
Table~\ref{joint-table-01} shows that DQAEM succeeds with the ratio of $97.4$ \% while EM succeeds with the ratio of $56.6$ \%, and that DQAEM is superior to EM.
In Table~\ref{ratio-table-12}, we show the ratios of success for DQAEM, EM, and DSAEM in parameter estimation.
This table also shows that DQAEM is superior to DSAEM.
\begin{table}[t]
\caption{Ratios of success for DQAEM, EM and DSAEM.}
\label{ratio-table-12}
\begin{center}
\begin{tabular}{|c|c|c|}
\hline
DQAEM  & EM    & DSAEM       \\ \hline
97.4 \%  & 56.6 \% & 77.8 \% \\
\hline
\end{tabular}
\end{center}
\end{table}

\section{Conclusion} \label{conc}

In this paper, we have proposed the deterministic quantum annealing expectation-maximization (DQAEM) algorithm for Gaussian mixture models (GMMs) to relax the problem of local optima of the expectation-maximization (EM) algorithm by introducing the mechanism of quantum fluctuations into EM.
Although we have limited our attention to GMMs in this paper to simplify the discussion, the derivation presented in this paper can be straightforwardly applied to any models which have discrete latent variables.
After formulating DQAEM, we have presented the theorem that guarantees its convergence.
We then have given numerical simulations to show its efficiency compared to EM and DSAEM.
It is expect that the combination of DQAEM and DSAEM gives better performance than DQAEM.
Finally, one of our future works is a Bayesian extension of this work.
In other words, we are going to propose a deterministic quantum annealing variational Bayes inference.

\bibliographystyle{IEEEtran}
\bibliography{IEEEabrv,paper-cdc-2016-99-01-bib}

\appendix

\subsection{Quantum annealing} \label{app-qa-01}

Here, we briefly introduce ``quantum" annealing (QA) to prepare for DQAEM.
First we consider the minimization problem of the Ising model.
That is,
\begin{align}
\min_{\{\sigma_z^{(i)}\}} H, \nonumber
\end{align}
where
\begin{align}
H &= - \sum_{i<j} J_{ij} \sigma_z^{(i)} \sigma_z^{(j)}, \label{Ising-01}
\end{align}
$\sigma_z^{(i)} = \pm 1$ for each $i$, and $J_{ij}$ is the coupling constant between spins at site $i$ and site $j$.
Note that this problem can describe many combinatorial problems such as the traveling salesman problem and the max-cut problem~\cite{Lucas01}.

In QA, we  quantize the Ising model~\eqref{Ising-01} by applying magnetic fields along the $x$ axis to the model and solve the Schr\"{o}dinger equation on this system while decreasing the magnetic fields.
Then the Hamiltonian of this system is given by
\begin{align}
\hat{H} = - \sum_{i<j} J_{ij} \hat{\sigma}_z^{(i)} \hat{\sigma}_z^{(j)} + \Gamma \sum_i \hat{\sigma}_x^{(i)}, \nonumber
\end{align}
where
\begin{align}
{\hat{\sigma}}_z^{(i)} &= \underbrace{\hat{I}_{2 \times 2} \otimes \dots \otimes \hat{I}_{2 \times 2}}_{i - 1} \otimes \hat{\sigma}_z \otimes \underbrace{\hat{I}_{2 \times 2} \otimes \dots \otimes \hat{I}_{2 \times 2}}_{N - i}, \nonumber \\
{\hat{\sigma}}_x^{(i)} &= \underbrace{\hat{I}_{2 \times 2} \otimes \dots \otimes \hat{I}_{2 \times 2}}_{i - 1} \otimes \hat{\sigma}_x \otimes \underbrace{\hat{I}_{2 \times 2} \otimes \dots \otimes \hat{I}_{2 \times 2}}_{N - i}, \nonumber
\end{align}
using
\begin{align}
\hat{I}_{2 \times 2} =
\begin{bmatrix}
1 & 0 \\
0 & 1
\end{bmatrix}, \
\hat{\sigma}_z =
\begin{bmatrix}
1 & 0 \\
0 & -1
\end{bmatrix}, \
\hat{\sigma}_x =
\begin{bmatrix}
0 & 1 \\
1 & 0
\end{bmatrix}, \nonumber
\end{align}
and $\Gamma$ represents the strength of the magnetic fields.
This is called the Transverse Ising model.
Thus the Schr\"{o}dinger equation that we solve in QA is given by
\begin{align}
i \hbar \frac{\partial}{\partial t} | \psi \rangle &= \hat{H} | \psi \rangle, \label{shrodinger-01}
\end{align}
where $i$ is the imaginary unit, $\hbar$ is the Dirac constant, and $| \psi \rangle$ is the ket vector.
The magnetic field $\Gamma$ is set to be large at the beginning of QA, and then $| \psi \rangle$ is initially equal or close to the eigenstate of $\sum_{i} \hat{\sigma}_x^{(i)}$.
During solving \eqref{shrodinger-01}, we gradually decrease $\Gamma$ and finally make $\Gamma$ go to zero.
Therefore, $| \psi \rangle$ gives a solution for the original Hamiltonian~\eqref{Ising-01}.
The efficiency of QA is discussed in Refs.~\cite{Kadowaki01, Farhi01, Morita01}.

\end{document}